\documentclass{bmvc2k}

\usepackage{times}
\usepackage{epsfig}
\usepackage{graphicx}
\usepackage{algorithm}
\usepackage[noend]{algpseudocode}
\usepackage{amsmath}
\usepackage{amssymb}
\usepackage{verbatim}
\usepackage{color}
\usepackage{multirow}
\usepackage{pifont}

\title{A Novel Self-Supervised Re-labeling Approach for Training with Noisy Labels}

\addauthor{Devraj Mandal*}{devrajm@iisc.ac.in}{1}
\addauthor{Shrisha Bharadwaj*}{shrishabharadwaj5@gmail.com}{1}
\addauthor{Soma Biswas}{somabiswas@iisc.ac.in}{1}

\addinstitution{
 Department of Electrical Engineering\\
 Indian Institute of Science\\
 Bangalore, India-560012.
}

\runninghead{Devraj}{Self-Supervised Re-labeling Method for Handling Noisy Labels}

\begin{document}

\maketitle

\begin{abstract}
	The major driving force behind the immense success of deep learning models is the availability of large datasets along with their clean labels.
	Unfortunately, this is very difficult to obtain, which has motivated research on the training of deep models in the presence of label noise and ways to avoid over-fitting on the noisy labels. 
	In this work, we build upon the seminal work in this area, {\textbf Co-teaching}  and propose a simple, yet efficient approach termed mCT-S2R (modified co-teaching with self-supervision and re-labeling) for this task.
	First, to deal with significant amount of noise in the labels, we propose to use self-supervision to generate robust features without using any labels.
	Next, using a parallel network architecture, an estimate of the clean labeled portion of the data is obtained.
	Finally, using this data, a portion of the estimated noisy labeled portion is re-labeled, before resuming the network training with the augmented data.
	Extensive experiments on three standard datasets show the effectiveness of the proposed framework.
	
\end{abstract}

\section{Introduction}

The success of deep learning models like Alexnet~\cite{alexnet}, VGG~\cite{vgg}, ResNet~\cite{resnet}, etc. for image classification tasks can be attributed to the availability of large, annotated datasets like ImageNet~\cite{imagenet}. 
But, obtaining clean annotations of large datasets is very expensive, and thus recent research has focused on learning from weakly supervised data, where the labels are often {\em noisy}, since they have been acquired from web searches \cite{websearch} or crowdsourcing \cite{crowd}. 
The presence of noisy labels can severely degrade the performance of the learned classifiers, since deep neural networks can even overfit on the noisy labels with sufficient training, due to their memorization capability \cite{memory1} \cite{memory2}.

Recently, several approaches have been proposed to handle label noise in the training data~\cite{memory1} \cite{coteaching} \cite{coteachingp} \cite{mentornet} \cite{decoupling} \cite{nal} \cite{loss_correct}. 
One direction to address this problem is to estimate the noise transition matrix itself by utilizing an additional softmax layer \cite{nal} for modeling the channel noise.
An alternative two-step approach along the same lines is proposed in \cite{loss_correct}. 
However, it is observed that estimation of the noise transition matrix is often hard and computationally expensive, especially when a large number of classes are present in the data \cite{coteaching}.    

A more recent and elegant direction to handle label noise is to utilise the concept of small-loss instances \cite{mentornet} \cite{ren2018learning} \cite{coteaching}.
Here, the algorithms estimate which samples are correctly labeled, and uses them to train the network subsequently. 
MentorNet \cite{mentornet} uses a pre-trained network (trained in a self-paced manner with a curriculum loss) to select clean labeled data to train the final classification model. 
Co-teaching \cite{coteaching} trains two networks in parallel and updates the weights of the networks using only the small loss samples. 
In addition, the gradients of the two networks are switched during the parameter update, which leads to better performance. 
It is observed~\cite{coteaching} that when the training continues for a long time, the two networks generally converge to the same state and start performing similarly, leading to the accumulation of errors.

Here, we propose a novel framework based on the Co-teaching approach~\cite{coteaching}, which also uses the concept of small-loss instances along with self-supervision and re-labeling, to significantly improve the training of deep networks with very noisy training data.
The proposed approach has four main steps.
First, to deal with significant amount of label noise, we utilize self-supervision as a pre-training step, so that the network can learn robust features without the use of any labels.
Second, we use a parallel network architecture similar to the one in \cite{coteaching} to get an estimate of the small loss samples.
In the third step, utilizing a portion of the small loss samples, the class means for all the categories are computed, which are then used to re-label the large loss samples. 
Finally, the network training is resumed by taking all the small loss samples along with a portion of the re-labeled large loss samples based on a confidence measure.
The proposed framework is termed as mCT-S2R (modified co-teaching with self-supervision and re-labeling).
The main contributions of this work are as follows: 
\begin{itemize}
	\item We develop a novel approach by utilizing self-supervision and re-labeling of the large loss data using the small loss samples for training deep networks under significant label noise. 
	\item Our framework uses two parallel networks only to determine the small loss instances. 
	Unlike \cite{coteaching} \cite{decoupling} which requires two networks for the entire training, the final training of the proposed framework after re-labeling proceeds using a single network. 
	This makes our model computationally much lighter.
	\item We propose using a self-supervised training paradigm like \cite{rotnet} to learn about the data distribution without using the labels, which helps to avoid overfitting on the noisy labels. 
	\item Extensive experiments on three benchmark datasets show the effectiveness of the proposed framework.
\end{itemize}


\section{Related Work}

Training deep neural networks in the presence of label noise is an active research area~\cite{decoupling}\cite{coteaching}\cite{coteachingp}\cite{noise_survey}.
Several directions have been proposed in the literature to address this problem.
Some works~\cite{noise2} \cite{noise_loss_fun1} have focussed on designing an appropriate loss function, which acts as a regularization term while training.
\cite{noise2} propose to increase the weight of the regularization term during the training paradigm so that the network focuses more on the cleaner examples with subsequent iterations.
An unbiased estimator to handle label noise is proposed in~\cite{noise_loss_fun2}, while a robust non-convex loss function is designed in~\cite{noise_loss_fun3} for this task.

Another research direction is to model the noise behavior itself~\cite{nal}\cite{loss_aerial}.
The work in \cite{loss_aerial} models the conditional probability of observing a wrong label, and considers the correct label to be a latent variable.
However, \cite{loss_aerial} assumes that the label-flip probability is known apriori, which requires extra knowledge about the training data.
This has been mitigated in~\cite{nal}, where an additional softmax layer is used to model the noise transition matrix.
Noise rate estimation is also addressed in the works of \cite{menon2015learning} \cite{liu2015classification} and \cite{sanderson2014class}.    

Most of the current approaches in this area utilize the concept of {\em small loss} instances \cite{mentornet}\cite{decoupling}\cite{coteaching}, where the network tries to estimate those samples which are probably correctly labeled, and ignore those samples which are probably noisy.
Due to ease of training these models and their generalization capability, these models have been well accepted in the machine learning community.
Our approach falls in this category, and we propose a novel framework mCT-S2R for this task.

There are some other related works~\cite{nigam2000analyzing}\cite{blum1998combining}\cite{ando2007two}\cite{zhu2005semi}, which assumes the availability of a small clean set of data in addition to the large, noisy or unlabeled dataset for training the classification model.
These approaches are beyond the scope of our work.


\section{Motivation}
\begin{figure}[t!]
	\centering
	\begin{center}            
		\includegraphics[width=0.60\linewidth, height=0.20\textheight]{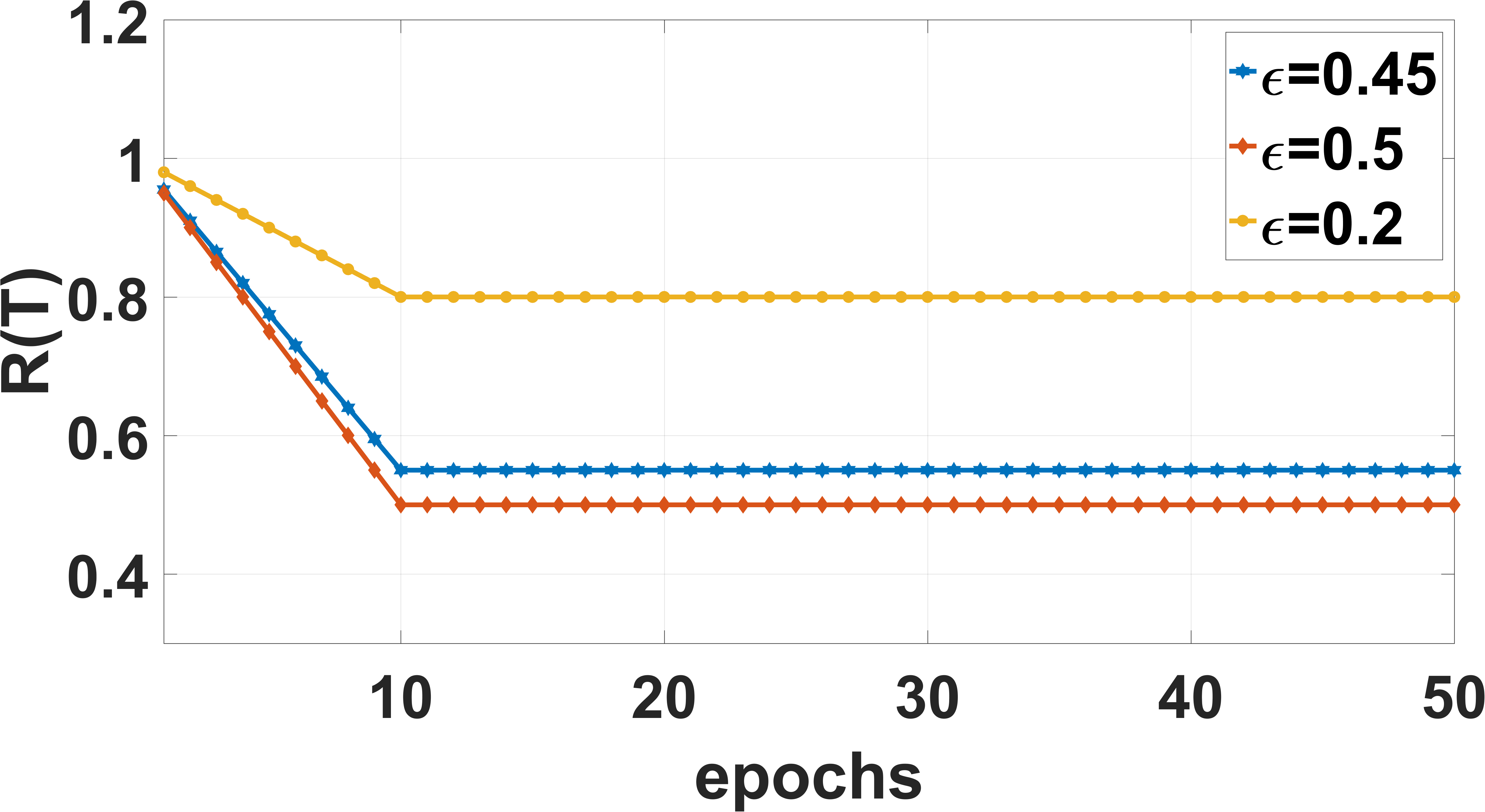}
	\end{center}
	\caption{Plot of $R(T)$ for different values of label corruption $\epsilon=\{0.45, 0.5, 0.2\}$. $R(T)$ is used ~\cite{coteaching} to control how many samples are used to update the network weights per epoch. The proposed approach augments the training set by adding a portion of the large loss samples after re-labeling them using the small loss instances.}
	\label{motivation}    
	\vspace{- 10 pt}
\end{figure}

To develop the proposed framework, first, we analyze the recently proposed methods which utilize the small loss instances~\cite{coteaching}.
These approaches typically use two network architecture, and each network updates the peer network weights based on the small loss samples.
The training procedure uses a factor $R(T)$ (at epoch $T$) to quantify the amount of small loss samples per mini-batch to be used to update the network weights.
$R(T)$ is directly related to the amount of label corruption $\epsilon$ ($\epsilon$ is either assumed to be known or can be estimated by using the strategies outlined in \cite{liu2015classification} \cite{yu2018efficient}), and is defined as $R(T) = 1 - \min \{\frac{T}{T_k} \epsilon, \epsilon\}$, where $T_k$ is the epoch after which $R(T)$ becomes constant.
Typical profiles of $R(T)$ for different $\epsilon$ is shown in Figure \ref{motivation}.
We observe that the model initially considers all the input data and slowly zones into the cleanly labeled data (guided by the concept of small loss instances) to train the model.
After completion of $T_k$ epochs (here $T_k=10$), the network weights are updated using $(1-\epsilon)$ portion of the training data (marked as small loss instances) and ignores the large loss samples under the assumption that they are wrongly labeled.
Though the small and large loss samples are selected at every epoch and mini-batch at random, the basic premise is to only focus on those samples whose labels are possibly correct.
This automatically leads to ignoring a large portion of the data samples for training the model.
In this work, we analyze whether the small loss samples can be used to correct the labels of some of the incorrectly labeled samples, which can then be utilized to better train the network.   

\begin{figure*}[t!]
	\centering
	\begin{center}            
		\includegraphics[width=0.9\linewidth, height=0.25\textheight]{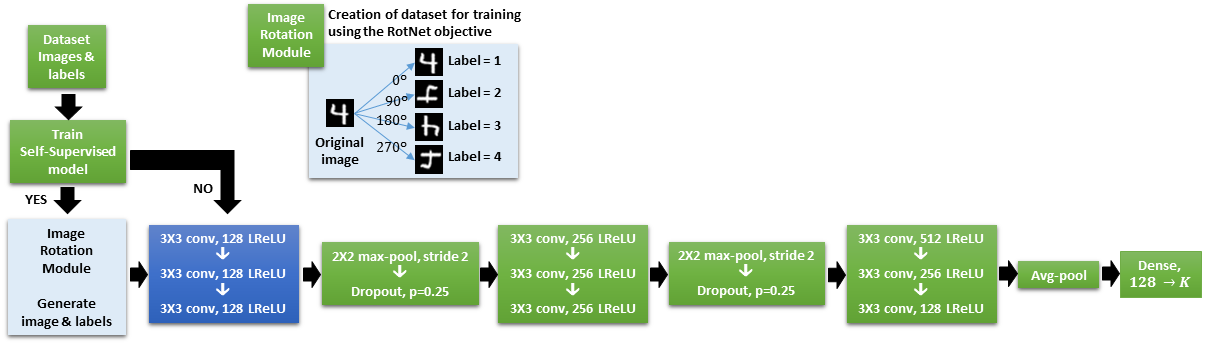}
	\end{center}
	\caption{
		The architecture of the network used for the pre-training task using the self-supervised objective.
		For the pretext task, each image is fed into the image rotation module (marked with light blue shade) to generate different rotated images and their corresponding labels (i.e., the degrees of rotation), and the network is trained using the standard cross-entropy loss. 
		For training the final noise-tolerant classification model, two networks of the same architectural design as shown are taken. The weights are randomly initialized and only the first 3 convolutional layer weights (marked with deep blue shade) are loaded with the weights from the pre-trained self-supervised model.
		The two parallel networks are trained using the images and their provided noisy labels.
		(Figure best viewed in color)}
	\label{main_fig}
	\vspace{- 10 pt}    
\end{figure*}

\section{Proposed Approach}

The proposed framework has four main steps: (1) pre-training the neural network in a self-supervised fashion to learn about the data distribution in an unsupervised manner, (2) training of two parallel networks following the same strategy as in \cite{decoupling} \cite{coteaching} to estimate the small loss instances, (3) re-labeling of the large loss samples using the small loss instances and, finally (4) final training of one network using an augmented set of the small loss samples and a portion of the re-labeled large loss samples.    
Psuedo-code of the proposed approach is provided in Algorithm 1.
Now, we describe the different steps in detail.

\subsection{Pre-training} 
To deal with significant amount of label noise, we first want to impart the knowledge of the training data distribution to the proposed network architecture without using the noisy labels. 
Recently, self-supervised models like \cite{rotnet}\cite{self_color}\cite{self_channel}\cite{self_context}\cite{self_jigsaw}\cite{self_move} have been used to learn high level convolutional network features in an unsupervised manner.
Features learned in a self-supervised way by solving a pretext task such as image colorization \cite{self_color}\cite{self_channel},  prediction of relative position of image patches \cite{self_context}\cite{self_jigsaw}, egomotion prediction by using motion between two frames \cite{self_move} and prediction of 2d rotation of the images \cite{rotnet} have helped to learn unsupervised features which are useful for further downstream computer vision tasks.

Inspired by this, we propose to first use a self-supervised training paradigm to learn a feature representation without using any label information. 
We have selected the RotNet model~\cite{rotnet} for our framework as shown in Figure~\ref{main_fig} (marked with light blue shade). 
Here, the pretext task is to correctly predict the rotation operation (such as image rotations of $0, 90, 180, 270$ degrees) which have been applied to the input image. 
The loss function used is the standard cross-entropy loss $\mathcal{L}^{ce}$ with a $4$-dim logit layer (for the $4$ degrees of orientation i.e., here $K=4$).

As mentioned, the proposed mCT-S2R is built on the Co-teaching approach~\cite{coteaching} and has the same base architecture (Figure \ref{main_fig}).
For seamless transfer of the learned weights from the self-supervised pretext task, we use the same architecture as in \cite{coteaching} with RotNet objective \cite{rotnet}.    
To train the model for the pretext task, we feed each image through the image rotation module (shown inset in Figure \ref{main_fig}) to generate the training data and its labels, i.e. the orientation degree.
This training is performed on the same dataset on which we need to train the classification model later i.e., the downstream task.
Once trained, this pre-trained network can be used for building the noise-robust classification model.

Based on the studies in \cite{rotnet}\cite{coteaching}\cite{decoupling}, we need to consider two important things: (1) the best way to transfer the knowledge learnt while solving the pretext task and (2) imparting randomness in the two parallel networks, so that they do not converge to the same model.
The network architecture for the two parallel networks is identical as shown in Figure \ref{main_fig}.    
Each network has a final dense layer of $K$-dim, which is randomly initialized, where, $K$ denotes the total number of training categories in the data.
The work in \cite{rotnet} has shown that the initial layers learn generic features as compared to the later layers, which are mostly tuned to the pretext task.
Thus to impart the knowledge of the data distribution, the weights of the first $3$ convolutional layers of both the networks in the proposed framework are initialized with the corresponding weights of the pre-trained self-supervised model.
The randomness of the rest of the layers prevents the two networks from converging to the same model during training.

\begin{algorithm}
	\caption{Proposed Framework - mCT-S2R}
	\label{algo_main}
	\begin{algorithmic}[t]
		\State \textbf{Input:} Network structure $\{p, q\}$, learning rate $\eta$, $\epsilon$, epoch $T_k$, $T_{update}$, $T_{\max}$, total number of mini-batches $N_{\max}$, number ($N$) of small loss instances to compute class means, training set $\mathcal{D}$, threshold value $\kappa$.
		\State \textbf{Output:} Trained noise robust classification model $p$.
		
		\vspace{2 pt}
		\State \textbf{\underline{Step 1: Pre-training}}
		\vspace{2 pt}
		\State 1. Randomly initialize one network and train it using the self-supervised objective loss function as in \cite{rotnet}.
		\State 2. Randomly initialize the model weights $w_p$ and $w_q$ for the two networks $p$ and $q$.
		\State 3. The initial three conv layer weights of both $p$ and $q$ are loaded with the weights of the pre-trained self-supervised model, while the remaining layers are randomly initialized.
		
		\vspace{2 pt}
		\State \textbf{\underline{Step 2: Compute small-loss instances}}
		\vspace{2 pt}
		\For{$T = 1,2, ..., T_{update}$}
		\State Shuffle the training set $\mathcal{D}$.
		\For{$N = 1, ..., N_{\max}$}
		\State Fetch mini-batch $\mathcal{D}^{mb}$ from $\mathcal{D}$.
		\State
		Sub-select a portion of the samples (as defined by $R(T)$) from $\mathcal{D}^{mb}$ to form the small loss set $\mathcal{D}^j$.
		\State Hence, $\mathcal{D}^j = \arg \min_{x: |x| \geq R(T)|\mathcal{D}^{mb}|} \mathcal{L}^{ce}(j, x) $
		\State Update $w_j = w_j - \eta \nabla \mathcal{L}^{ce} (j, \mathcal{D}^{\bar{j}})$.
		\State $(j = \{p,q \}$, if $j=p, \bar{j} = q)$
		\EndFor
		\State Update $R(T) = 1 - \min \left( \frac{T}{T_k} \epsilon, \epsilon \right)$.
		\EndFor
		
		\vspace{2 pt}
		\State \textbf{\underline{Step 3: Re-labeling large-loss samples}}
		\vspace{2 pt}
		\State 1. Pick one network randomly (say $p$). Store the sorted (in ascending order) small and large loss indices at $T_{update}$ epoch as $\{S^p, L^p\}$. 
		The small loss samples are denoted as as $\mathcal{D}^s$. 
		\State 2. Compute the class-wise means using the the top-N small loss instances (\textbf{Algorithm2}).
		\State 3. Based on distance of the large loss samples from the means, generate the re-labeled targets $\mathcal{Y}^r$ and their confidence of re-labeling $\mathcal{C}$ (\textbf{Algorithm3}).
		\State 5. Select the large loss samples whose confidence $c \geq \kappa$ to form $\mathcal{D}^r$.
		\State 6. Construct augmented training set $\mathcal{D}^{aug} = \mathcal{D}^s \cup \mathcal{D}^r$.
		
		\vspace{2 pt}
		\State \textbf{\underline{Step 4: Final training with augmented data}}
		\vspace{2 pt}            
		\For{$T = T_{update}+1,T_{update}+2, ..., T_{\max}$}
		\State Shuffle the training set $\mathcal{D}^{aug}$.
		\For{$N = 1, ..., N_{\max}$}
		\State Fetch mini-batch $\mathcal{D}^{mb}$ from $\mathcal{D}^{aug}$.
		\State Obtain $\mathcal{D}^p = \arg \min_{x: |x| \geq R(T)|\mathcal{D}^{mb}|} \mathcal{L}^{ce}(p, x) $
		\State Update $w_p = w_p - \eta \nabla \mathcal{L}^{ce} (p, \mathcal{D}^p)$  .      
		\EndFor
		\State Update $R(T) = 1 - \min \left( \frac{T}{T_k} \epsilon, \epsilon \right)$.
		\EndFor        
	\end{algorithmic}
\end{algorithm}    

\subsection{Computing the small-loss instances} 
We train two parallel networks, each one having the architecture as in Figure \ref{main_fig} simultaneously as in \cite{decoupling} \cite{coteaching} by feeding the original image data and the corresponding noisy labels.
Let us denote the two networks as $p$ and $q$ and their learnable weights as $w_p$ and $w_q$. For each network, we denote the output from the final, fully connected (fc) layer logits to be $\{u^p, u^q\}$ of dimension $K$, and the penultimate fc layer features to be $\{f^p, f^q\}$ ($128$-dim). 
The algorithm proceeds as in \cite{coteaching}, where, in each mini-batch, the network computes the loss over the available samples in the mini-batch, sorts the loss in ascending order, and selects a portion of the samples to form the small loss instance set $\{\mathcal{D}^p, \mathcal{D}^q\}$ based on the value of $R(T)$. 
The parameters $w_p$ and $w_q$ are updated by using the information from the peer network i.e., $\mathcal{D}^q$ and $\mathcal{D}^p$. The loss to train the network is the standard cross-entropy loss $\mathcal{L}^{ce} = - \log \left( e^{u[i]} / \sum_{k=1}^{K} e^{u[k]} \right)$, where, $K$ is the number of categories and $i$ is the correct class index.

In~\cite{coteaching}, the two networks are trained until the final number of iterations $T_{\max}$.
During testing, one of the learned networks (either $p$ or $q$) is used for classification.
In contrast, we train the two parallel networks for an initial $T_{update}$ number of steps and then collect all the small loss ($\{S^p,S^q\}$) and large loss ($\{L^p,L^q\}$) indices set based on the value of $R(T)$.
Unlike~\cite{coteaching}, we use the two networks only to get an estimate of the small and large loss samples.
For subsequent training, we choose one of the networks, thus reducing the computational requirement significantly.

\subsection{Re-labeling the large loss samples} 
Let us consider, that we choose the network $p$ and its index set $\{S^p, L^p\}$.
The set of all small loss samples is denoted as $\mathcal{D}^s$.
In this step, first, all the training data is passed through the network $p$ and their features ($\mathcal{F} = \{f^p\}$) are extracted.
The indices $S^p$ and $L^p$ are used to compute the feature sets for the small loss and large loss samples. 
The per class mean feature representation is then computed using the small loss sample features to form the mean set $\mathcal{M} = \{\mu_1, \mu_2, ..., \mu_K\}$.
Instead of using all the samples of each class to compute the means, we use only the top-$N$ small loss samples per category, i.e the samples with the lowest loss.
This step of feature extraction and class-wise mean computation is illustrated in Algorithm 2.

Next, we use the class-wise means $\mathcal{M}$ and the extracted features of the large loss samples to generate their pseudo-labels $\mathcal{Y}^r$ and confidence values $\mathcal{C}$. 
The pseudo labels are generated by computing the distance of each large loss sample feature from $\mathcal{M}$ and converting it into a softmax probability score. 
The index of the highest softmax value gives the pseudo-label $y^r$, while the highest value is stored as its confidence measure value $c$. 
The steps are outlined in Algorithm 3.

%

\begin{algorithm}[t!]
	\caption{Feature Extraction \& Mean Set Computation}\label{algo_get_center}
	\begin{algorithmic}[t]
		\State \textbf{Input:} Network $p$, training data $\mathcal{D}$, sorted small \& large loss index set $\{S^p, L^p\}$, $N$: number of samples for computing class means.
		\State \textbf{Output:} Features of large loss samples $\mathcal{F}^L$, class-wise means $\mathcal{M} = \{\mu_1,...,\mu_K\}$ for $K$ categories
		\State 1. Pass data $\mathcal{D}$ through network $p$ and generate the extracted features.
		\State 2. Use index set $S^p$ and $L^p$ to get the features for large loss and small loss samples as $\mathcal{F}^L$ and $\mathcal{F}^S$.
		\State 3. Use $N$ small loss samples per category to compute its mean $\mu_k$ for class $k \in K$.
	\end{algorithmic}
	\vspace{- 4 pt}
\end{algorithm}
\begin{algorithm}[t!]
	\caption{Re-labeling \& Confidence Computation}\label{relabel}
	\begin{algorithmic}[t]
		\State \textbf{Input:} Feature set of large loss samples $\mathcal{F}^L$, class-wise means $\mathcal{M} = \{\mu_1,...,\mu_K\}$
		\State \textbf{Output:} Re-labeled targets $\mathcal{Y}^r=\{y^r\}$, Confidence values $\mathcal{C}=\{c\}$
		\For{$i = 1, ..., |\mathcal{F}^L|$}
		\State Compute distance $d$ of the $i^{th}$ sample from $\mathcal{M}$.
		\State Generate similarity value $\hat{d}$ by passing $(-d)$ through softmax function. (A smaller distance denotes a higher similarity.)
		\State Set pseudo label as: $y^r = \arg \max_k \hat{d}[k]$
		\State Set confidence of re-labeling as: $c = \max_k \hat{d}[k]$
		\EndFor
	\end{algorithmic}
	\vspace{- 3 pt}
\end{algorithm}

\subsection{Final training with augmented data}
Once we obtain the re-labeled samples and their confidence scores, we sub-select a portion of the large loss samples, whose confidence $c \geq \kappa$.
Let us denote this set by $\mathcal{D}^r$.
Finally, we construct the augmented training set as $\mathcal{D}^{aug} = \mathcal{D}^s \cup \mathcal{D}^r$.
We use the original provided labels for the small loss instances i.e., $\mathcal{D}^s$ and the predicted labels for the large loss samples i.e., $\mathcal{D}^r$ to further train the chosen network (here $p$).

\section{Experimental Evaluation}
Here, we describe in details the extensive experiments performed to evaluate the effectiveness of the proposed framework.
\subsection{Dataset details and noise models}

We conduct experiments on three standard benchmark datasets used for evaluating models trained with noisy labels, namely MNIST~\cite{mnist}, CIFAR-10~\cite{cifar} and CIFAR-100~\cite{cifar}. 
MNIST~\cite{mnist} dataset contains grayscale images of handwritten digits of size $28 \times 28$ spread over $10$ different categories.
This dataset has $60,000$ training and $10,000$ testing examples. 
CIFAR-10 \cite{cifar} and CIFAR-100 \cite{cifar} contains a huge collection of $32 \times 32$ size color images which has been divided into a training:testing split of $50,000:10,000$ examples. 
CIFAR-100 (CIFAR-10) data contains $100$ ($10$) different categories, with each category having $600$ ($6000$) images per class.

We follow the standard protocol as in \cite{loss_correct}\cite{noise2} to introduce label noise into the datasets.
Specifically,  the labels are corrupted using a noise transition matrix $P$, where $P_{ij} = \text{Prob} (\hat{l}=j | l=i)$, i.e.,  probability that the label of an example actually belonging to category $i$ ($l=i$) has been corrupted to belong to category $j$ ($\hat{l}=j$). 
We have used two different noise models: (1) symmetry flipping \cite{symmetry} (symmetry) and (2) pair flipping (pair). 
In symmetry flipping, the probability of an example being correctly labeled under the noisy setting is $(1-\epsilon)$, while the probability of being corrupted (i.e. $\epsilon$) is equally distributed over the rest of the categories. 
In pair flipping, the probability of being corrupted is distributed over a single category.
Studies in \cite{coteaching} have shown that it is typically harder to train classification networks under the pairflip noise model as compared to the symmetry noise model which is further validated by the performance of the algorithms under those settings.

We evaluate our framework under extreme noisy setting \cite{coteaching}, with $\epsilon = 0.45$ (for pairflip) and $\epsilon = 0.5$ (for symmetry), where almost half of the training examples have been corrupted and have noisy labels. 
Studies in \cite{coteaching} have shown that algorithms need additional information to handle more label noise i.e., $\epsilon > 0.5$ in the training data.

\begin{table}[b!]
	\small
	\centering
	\caption{Average test accuracy on the MNIST, CIFAR-10 and CIFAR-100 datasets over the last ten epochs in the supervised setting (with $100\%$ clean labels).}
	\renewcommand{\arraystretch}{1.2}
	\setlength{\tabcolsep}{12 pt}
	\begin{tabular}{|c|c|c|c|}
		\hline
		& MNIST & CIFAR-10 & CIFAR-100 \\ \hline
		Acc & 99.60 & 89.34 & 63.41 \\ \hline
	\end{tabular}
	\label{supervised}
	\vspace{- 10 pt}
\end{table}    

\begin{table*}[t!]
	\tiny
	\centering
	\caption{Average test accuracy on the MNIST dataset over the last ten epochs. The results of our approach without (and with) self-supervision pre-training is denoted as mCT-R and mCT-S2R respectively.}
	\renewcommand{\arraystretch}{1.1}
	\setlength{\tabcolsep}{7 pt}
	\begin{tabular}{|c|c|c|c|c|c|c|c|c|c|}
		\hline
		Flipping-Rate & Standard & Bootstrap & S-model & F-correction & Decoupling & MentorNet & Co-teaching & mCT-R & mCT-S2R \\ \hline
		Pair-45$\%$ & \begin{tabular}[c]{@{}c@{}}56.52\\ $\pm$ 0.55\end{tabular} & \begin{tabular}[c]{@{}c@{}}57.23\\ $\pm$ 0.73\end{tabular} & \begin{tabular}[c]{@{}c@{}}56.88\\ $\pm$ 0.32\end{tabular} & \begin{tabular}[c]{@{}c@{}}0.24\\ $\pm$ 0.03\end{tabular} & \begin{tabular}[c]{@{}c@{}}58.03\\ $\pm$ 0.07\end{tabular} & \begin{tabular}[c]{@{}c@{}}80.88\\ $\pm$ 4.45\end{tabular} & \begin{tabular}[c]{@{}c@{}}87.63\\ $\pm$ 0.21\end{tabular} & \begin{tabular}[c]{@{}c@{}}{\bf 95.52}\\ {\bf $\pm$ 0.04} \end{tabular} & \begin{tabular}[c]{@{}c@{}}94.72\\ $\pm$ 0.02\end{tabular} \\ \hline
		Symmetry-50$\%$ & \begin{tabular}[c]{@{}c@{}}66.05\\ $\pm$ 0.61\end{tabular} & \begin{tabular}[c]{@{}c@{}}67.55\\ $\pm$ 0.53\end{tabular} & \begin{tabular}[c]{@{}c@{}}62.29\\ $\pm$ 0.46\end{tabular} & \begin{tabular}[c]{@{}c@{}}79.61\\ $\pm$ 1.96\end{tabular} & \begin{tabular}[c]{@{}c@{}}81.15\\ $\pm$ 0.03\end{tabular} & \begin{tabular}[c]{@{}c@{}}90.05\\ $\pm$ 0.30\end{tabular} & \begin{tabular}[c]{@{}c@{}}91.32\\ $\pm$ 0.06\end{tabular} & \begin{tabular}[c]{@{}c@{}}97.79\\ $\pm$ 0.02\end{tabular} & \begin{tabular}[c]{@{}c@{}}{\bf 97.82}\\ {\bf $\pm$ 0.02} \end{tabular} \\ \hline
		Symmetry-20$\%$ & \begin{tabular}[c]{@{}c@{}}94.05\\ $\pm$ 0.16\end{tabular} & \begin{tabular}[c]{@{}c@{}}94.40\\ $\pm$ 0.26\end{tabular} & \begin{tabular}[c]{@{}c@{}}98.31\\ $\pm$ 0.11\end{tabular} & \begin{tabular}[c]{@{}c@{}}98.80\\ $\pm$ 0.12\end{tabular} & \begin{tabular}[c]{@{}c@{}}95.70\\ $\pm$ 0.02\end{tabular} & \begin{tabular}[c]{@{}c@{}}96.70\\ $\pm$ 0.22\end{tabular} & \begin{tabular}[c]{@{}c@{}}97.25\\ $\pm$ 0.03\end{tabular} & \begin{tabular}[c]{@{}c@{}} {\bf 99.05} \\ {\bf $\pm$ 0.03} \end{tabular} & \begin{tabular}[c]{@{}c@{}}98.86\\ $\pm$ 0.01\end{tabular} \\ \hline
	\end{tabular}
	\label{table_mnist}
	\vspace{- 10 pt}
\end{table*}


\subsection{Network structure and implementation details}    
We implement the proposed framework in PyTorch (version 0.4.1) \cite{pytorch} on a NVIDIA GTX 1080 Ti GPU using the architecture shown in Figure \ref{main_fig}.
The model is trained with Adam optimizer using momentum of $0.9$ and a learning rate of $0.001$.
First, the model is trained using self-supervision based on the RotNet objective for $25$ epochs.
For handling large amounts of label noise, the first three convolutional layers of the two parallel networks are initialized with pre-trained weights of the self-supervised model and the remaining layers are initialized with random weights.
We use a batch size of $128$ and train for $200$ epochs, and report the average test accuracy over the last ten epochs.
We follow the same protocol as in \cite{coteaching} and fix the other parameters of our model as $R(T)=1-\epsilon. \min (\frac{T}{T_k}, 1), T_k=10, T_{max}=200$.
We set $T_{update}=30$ and the softmax confidence measure $\kappa=0.90$ for all our experiments.
We also perform ablation studies to analyze the performance of our algorithm with the change in $T_{update}$ and $\kappa$ and report the results later.

We compare the proposed approach against the state-of-art approaches: Bootstrap \cite{noise2}, S-model \cite{nal}, F-correction \cite{loss_correct}, Decoupling \cite{decoupling}, MentorNet \cite{mentornet} and Co-teaching \cite{coteaching}. 
The results of all the compared methods are taken from the published results in \cite{coteaching}.    
We also compute the classification accuracy of the proposed mCT-S2R on the three datasets using all the clean labels.
The results in Table \ref{supervised} represents the upper-bound of classification accuracy that can possibly be achieved by our network.

\begin{figure}[t!]
	\centering
	\begin{center}            
		\includegraphics[width=0.8\linewidth, height=0.25\textheight]{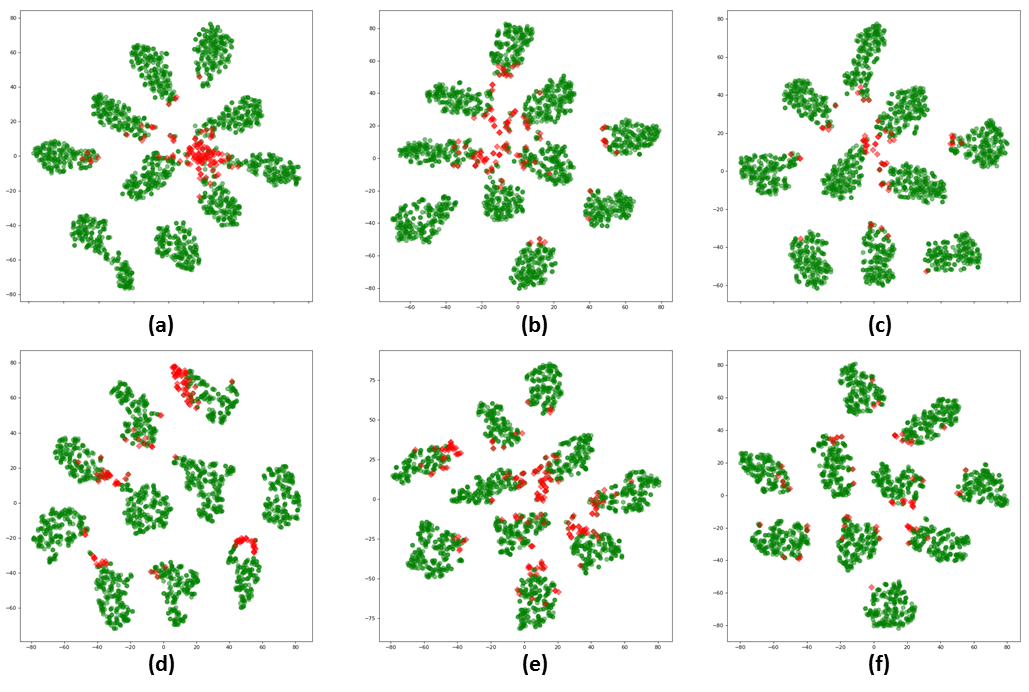}
	\end{center}
	\caption{T-sne plots of the large loss samples for MNIST. 
		The small loss samples are used to re-label the large loss instances and the correctly (wrongly) re-labeled data are marked with green (red) respectively. (a, b, c) and (d, e, f) denote the results of mCT-R and mCT-S2R. 
		(Left to right): The three columns correspond to the noise models: pairflip-$45\%$, symmetric-$50\%$ and symmetric-$20\%$.
		(Figure best viewed in color)}
	\label{tsne_mnist}    
	\vspace{- 10 pt}
\end{figure}

\subsection{Evaluation on MNIST dataset}    
The results of the proposed mCT-S2R under different noise models on the MNIST dataset is reported in Table \ref{table_mnist}. 
We also report the results without self-supervision (denoted by mCT-R) to analyze the importance of the two major components of the proposed framework, namely self-supervision and re-labeling.
We observe the following:
(1) For all the approaches, the performance for the pairflip noise model is lower as compared to symmetry flip;
(2) As expected, the performance of all the algorithms for $20\%$ symmetric noise is better than for $50\%$ noise;
(3) The proposed approach significantly outperforms the state-of-the-art models, especially for pairflip noise;
(4) The performance of mCT-S2R for the symmetric case is close to the best possible value (shown in Table \ref{supervised}).
This might be because MNIST is an easier dataset compared to the others since the data for a particular category in MNIST has comparatively fewer variations.
(5) The results of mCT-S2R and mCT-R are very similar, signifying that self-supervision is not particularly beneficial for this dataset.

Now, we analyze the usefulness of re-labeling in the proposed framework.
We train the two parallel networks till $T_{update}$, then randomly select one of them, and use the small loss instances to predict the labels of the large loss instances. 
The t-sne \cite{tsne} plots of the large loss sample features are shown in Figure \ref{tsne_mnist}.
The large loss samples which are correctly re-labeled are marked with {\em green}, while those which are wrongly re-labeled are marked with {\em red}.
The top (bottom) row denotes the results without (with) self-supervised pre-training.    
We observe that most of the large loss samples are getting correctly re-labeled in both the cases.
Based on a softmax confidence threshold, our framework selects a portion of the re-labeled samples (preferably from the green marked ones) to train the network further.
This extra set of correctly re-labeled samples helps our approach to achieve better classification accuracy.

\begin{table*}[t!]
	\tiny
	\centering
	\caption{Average test accuracy on the CIFAR-10 dataset over the last ten epochs. The results of our approach without (and with) self-supervision pre-training is denoted as mCT-R and mCT-S2R respectively.}
	\renewcommand{\arraystretch}{1.1}
	\setlength{\tabcolsep}{7 pt}
	\begin{tabular}{|c|c|c|c|c|c|c|c|c|c|}
		\hline
		Flipping-Rate & Standard & Bootstrap & S-model & F-correction & Decoupling & MentorNet & Co-teaching & mCT-R & mCT-S2R \\ \hline
		Pair-45$\%$ & \begin{tabular}[c]{@{}c@{}}49.50\\ $\pm$ 0.42\end{tabular} & \begin{tabular}[c]{@{}c@{}}50.05\\ $\pm$ 0.30\end{tabular} & \begin{tabular}[c]{@{}c@{}}48.21\\ $\pm$ 0.55\end{tabular} & \begin{tabular}[c]{@{}c@{}}6.61\\ $\pm$ 1.12\end{tabular} & \begin{tabular}[c]{@{}c@{}}48.80\\ $\pm$ 0.04\end{tabular} & \begin{tabular}[c]{@{}c@{}}58.14\\ $\pm$ 0.38\end{tabular} & \begin{tabular}[c]{@{}c@{}}72.62\\ $\pm$ 0.15\end{tabular} & \begin{tabular}[c]{@{}c@{}}78.09\\ $\pm$ 0.11\end{tabular} & \begin{tabular}[c]{@{}c@{}} {\bf 80.58}\\ {\bf $\pm$ 0.54} \end{tabular} \\ \hline
		Symmetry-50$\%$ & \begin{tabular}[c]{@{}c@{}}48.87\\ $\pm$ 0.52\end{tabular} & \begin{tabular}[c]{@{}c@{}}50.66\\ $\pm$ 0.56\end{tabular} & \begin{tabular}[c]{@{}c@{}}46.15\\ $\pm$ 0.76\end{tabular} & \begin{tabular}[c]{@{}c@{}}59.83\\ $\pm$ 0.17\end{tabular} & \begin{tabular}[c]{@{}c@{}}51.49\\ $\pm$ 0.08\end{tabular} & \begin{tabular}[c]{@{}c@{}}71.10\\ $\pm$ 0.48\end{tabular} & \begin{tabular}[c]{@{}c@{}}74.02\\ $\pm$ 0.04\end{tabular} & \begin{tabular}[c]{@{}c@{}}77.69\\ $\pm$ 0.24\end{tabular} & \begin{tabular}[c]{@{}c@{}}{\bf 81.23}\\ {\bf $\pm$ 0.07}\end{tabular} \\ \hline
		Symmetry-20$\%$ & \begin{tabular}[c]{@{}c@{}}76.25\\ $\pm$ 0.28\end{tabular} & \begin{tabular}[c]{@{}c@{}}77.01\\ $\pm$ 0.29\end{tabular} & \begin{tabular}[c]{@{}c@{}}76.84\\ $\pm$ 0.66\end{tabular} & \begin{tabular}[c]{@{}c@{}}84.55\\ $\pm$ 0.16\end{tabular} & \begin{tabular}[c]{@{}c@{}}80.44\\ $\pm$ 0.05\end{tabular} & \begin{tabular}[c]{@{}c@{}}80.76\\ $\pm$ 0.36\end{tabular} & \begin{tabular}[c]{@{}c@{}}82.32\\ $\pm$ 0.07\end{tabular} & \begin{tabular}[c]{@{}c@{}}84.89 \\ $\pm$ 0.09\end{tabular} & \begin{tabular}[c]{@{}c@{}}{\bf 87.21}\\ {\bf $\pm$ 0.04}\end{tabular} \\ \hline
	\end{tabular}
	\label{table_cifar10}
\end{table*}    
\begin{table*}[t!]
	\tiny
	\centering
	\caption{Average test accuracy on the CIFAR-100 dataset over the last ten epochs. The results of our approach without (and with) self-supervision pre-training is denoted as mCT-R and mCT-S2R respectively.}
	\renewcommand{\arraystretch}{1.1}
	\setlength{\tabcolsep}{7 pt}
	\begin{tabular}{|c|c|c|c|c|c|c|c|c|c|}
		\hline
		Flipping-Rate & Standard & Bootstrap & S-model & F-correction & Decoupling & MentorNet & Co-teaching & mCT-R & mCT-S2R \\ \hline
		Pair-45$\%$ & \begin{tabular}[c]{@{}c@{}}31.99\\ $\pm$ 0.64\end{tabular} & \begin{tabular}[c]{@{}c@{}}32.07\\ $\pm$ 0.30\end{tabular} & \begin{tabular}[c]{@{}c@{}}21.79\\ $\pm$ 0.86\end{tabular} & \begin{tabular}[c]{@{}c@{}}1.60\\ $\pm$ 0.04\end{tabular} & \begin{tabular}[c]{@{}c@{}}26.05\\ $\pm$ 0.03\end{tabular} & \begin{tabular}[c]{@{}c@{}}31.60\\ $\pm$ 0.51\end{tabular} & \begin{tabular}[c]{@{}c@{}}34.81\\ $\pm$ 0.07\end{tabular} & \begin{tabular}[c]{@{}c@{}}31.80\\ $\pm$ 0.11\end{tabular} & \begin{tabular}[c]{@{}c@{}}{\bf 38.67} \\ {\bf $\pm$ 0.08} \end{tabular} \\ \hline
		Symmetry-50$\%$ & \begin{tabular}[c]{@{}c@{}}25.21\\ $\pm$ 0.64\end{tabular} & \begin{tabular}[c]{@{}c@{}}21.98\\ $\pm$ 6.36\end{tabular} & \begin{tabular}[c]{@{}c@{}}18.93\\ $\pm$ 0.39\end{tabular} & \begin{tabular}[c]{@{}c@{}}41.04\\ $\pm$ 0.07\end{tabular} & \begin{tabular}[c]{@{}c@{}}25.80\\ $\pm$ 0.04\end{tabular} & \begin{tabular}[c]{@{}c@{}}39.00\\ $\pm$ 1.00\end{tabular} & \begin{tabular}[c]{@{}c@{}}41.37\\ $\pm$ 0.08\end{tabular} & \begin{tabular}[c]{@{}c@{}}41.46\\ $\pm$ 0.06\end{tabular} & \begin{tabular}[c]{@{}c@{}}{\bf 52.87} \\ {\bf $\pm$ 0.16} \end{tabular} \\ \hline
		Symmetry-20$\%$ & \begin{tabular}[c]{@{}c@{}}47.55\\ $\pm$ 0.47\end{tabular} & \begin{tabular}[c]{@{}c@{}}47.00\\ $\pm$ 0.54\end{tabular} & \begin{tabular}[c]{@{}c@{}}41.51\\ $\pm$ 0.60\end{tabular} & \begin{tabular}[c]{@{}c@{}} {\bf 61.87}\\ {$\pm$ \bf 0.21} \end{tabular} & \begin{tabular}[c]{@{}c@{}}44.52\\ $\pm$ 0.04\end{tabular} & \begin{tabular}[c]{@{}c@{}}52.13\\ $\pm$ 0.40\end{tabular} & \begin{tabular}[c]{@{}c@{}}54.23\\ $\pm$ 0.08\end{tabular} & \begin{tabular}[c]{@{}c@{}}56.29 \\ $\pm$ 0.19\end{tabular} & \begin{tabular}[c]{@{}c@{}}{ 60.49}\\ {$\pm$ 0.36} \end{tabular} \\ \hline
	\end{tabular}
	\label{table_cifar100}
\end{table*}

\subsection{Evaluation on CIFAR-10 dataset}        
We report the results of our approach on CIFAR-10 dataset in Table \ref{table_cifar10}. 
Even for this data, we observe similar trends in the results as with MNIST.
But for CIFAR-10, the results of mCT-S2R are significantly better than mCT-R, which clearly shows the effectiveness of self-supervised pre-training for more difficult data.

\begin{figure}[t!]
	\centering
	\begin{center}            
		\includegraphics[width=0.8\linewidth, height=0.25\textheight]{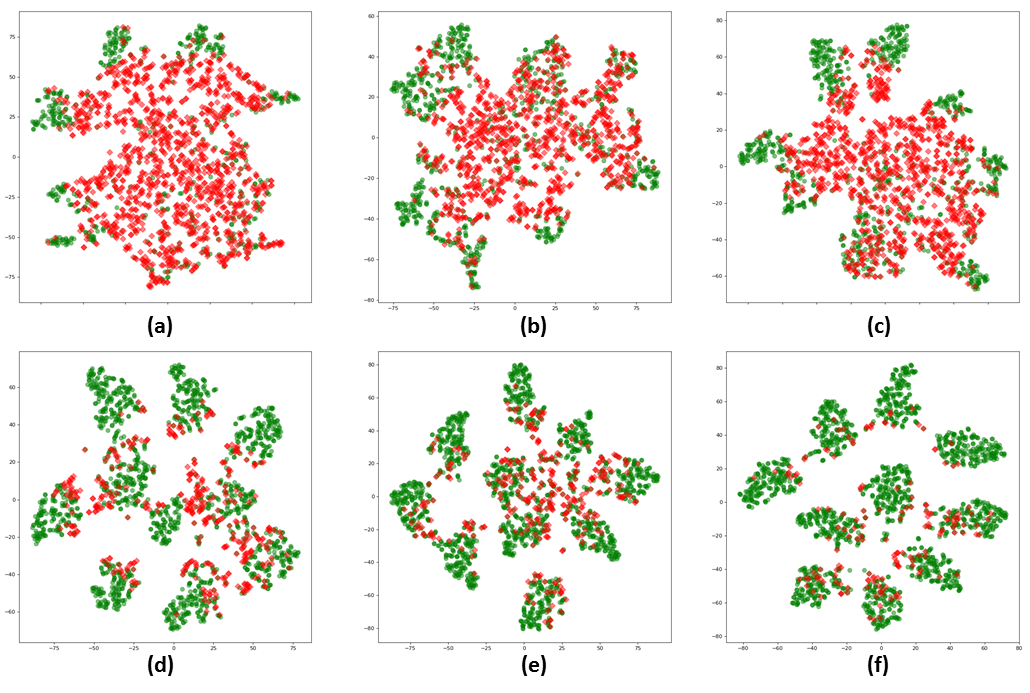}
	\end{center}
	\caption{T-sne plots of the large loss samples for CIFAR-10. The small loss samples are used to re-label the large loss instances and the correctly (wrongly) re-labeled data are marked with green (red) respectively. (a, b, c) and (d, e, f) denote the results of mCT-R and mCT-S2R.
		(Left to right): The three columns correspond to the noise models: pairflip-$45\%$, symmetric-$50\%$ and symmetric-$20\%$.
		(Figure best viewed in color)}
	\label{tsne_cifar10}
	\vspace{- 10 pt}    
\end{figure}

The t-sne plots for the different losses in Figure \ref{tsne_cifar10} (top row) clearly indicate that for the models trained without self-supervision, distinct clusters of data have not been formed and the number of wrongly marked samples (marked in "red") is quite high.
Thus, the re-labeled samples will potentially have many wrongly assigned labels.
In the bottom row, self-supervision pre-training clearly helps in forming distinct ($10$) clusters of data leading to a larger number of correctly re-labeled samples (marked in "green").
These additional correctly labeled data helps the proposed framework to achieve significantly better classification performance.
\subsection{Evaluation on CIFAR-100 dataset}
\begin{figure}[t!]
	\centering
	\begin{center}            
		\includegraphics[width=0.8\linewidth, height=0.25\textheight]{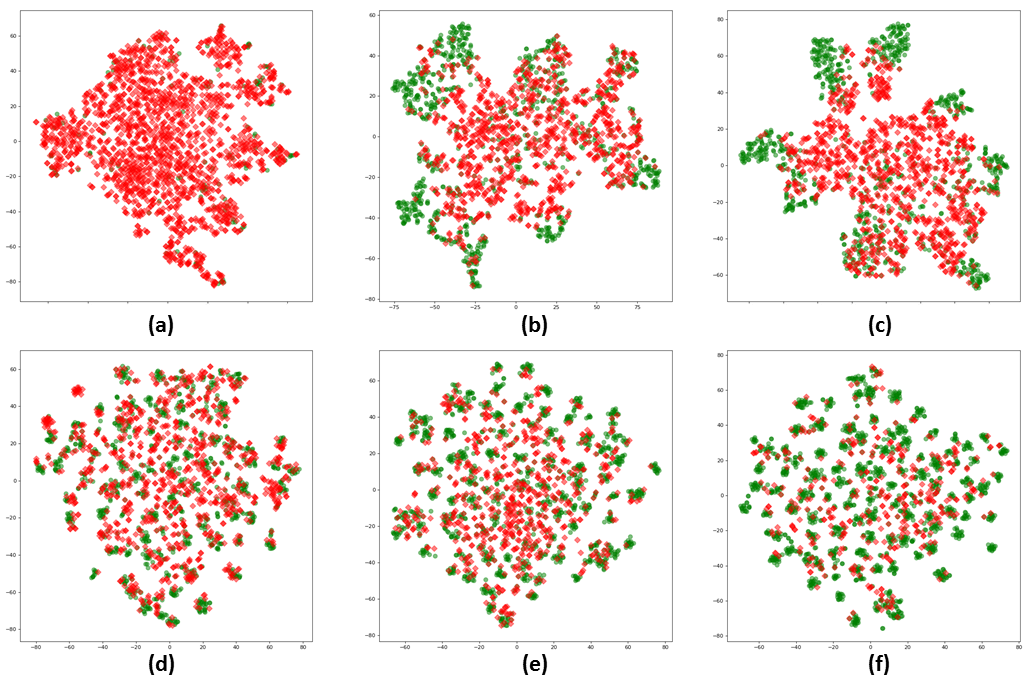}
	\end{center}
	\caption{T-sne plots of the large loss samples for CIFAR-100. The small loss samples are used to re-label the large loss instances and the correctly (wrongly) re-labeled data are marked with green (red) respectively. (a, b, c), (d, e, f) denote the results of mCT-R and mCT-S2R. 
		(Left to right): The three columns correspond to noise models: pairflip-$45\%$, symmetric-$50\%$ and symmetric-$20\%$.
		(Figure best viewed in color)}
	\label{tsne_cifar100}
\end{figure}

We report the results of our algorithm on the CIFAR-100 dataset in Table \ref{table_cifar100}. 
This is the hardest dataset of all the three due to the presence of $100$ different categories, which increases the possibility of incorrect re-labeling of the large loss samples.
Indeed, we observe that for pairflip-$45\%$, mCT-R shows poorer performance than \cite{coteaching}, possibly due to incorrect re-labeling of the large loss samples.
For this dataset, mCT-S2R shows significant improvements as compared to mCT-R.    
From the t-sne plots for the different losses in Figure \ref{tsne_cifar100}, we clearly observe how self-supervised pre-training immensely helps to correctly re-label the samples leading to the improved performance of mCT-S2R.


\subsection{Ablation studies}

Here, we provide an additional analysis of the proposed framework.
The analysis of why self-supervised pre-training helps to improve the performance of the network is shown in Figure \ref{tsne_mnist}, \ref{tsne_cifar10}, \ref{tsne_cifar100}.    
Further analysis for different choice of the hyper-parameters is provided below. 

\textbf{Choice of $T_{update}$:} 
The $T_{update}$ parameter controls the epoch after which we re-label the large loss samples using the small loss instances. 
We analyze the performance of the proposed mCT-S2R using different values of $T_{update}=\{10,20,30\}$ and report the results in Table \ref{ablation2} on MNIST and CIFAR-100 datasets.
We observe that the results are quite consistent, which implies that the proposed framework performs well for a wide range of hyper-parameter $T_{update}$ value.
Having a smaller value of $T_{update}$ is advantageous for our algorithm since after re-labeling, we can train only a single network which is computationally much lighter.

\begin{table}[h!]
	\small
	\centering
	\caption{Average test accuracy (over the last ten epochs) on MNIST and CIFAR-100 datasets for different values of $T_{update}$.}
	\renewcommand{\arraystretch}{1.1}
	\setlength{\tabcolsep}{5.0 pt}
	\begin{tabular}{|c|c|c|c|c|c|c|}
		\hline
		MNIST & \multicolumn{3}{c|}{Symmetry-$50\%$} & \multicolumn{3}{c|}{Pair-$45\%$} \\ \hline
		$T_{update}$ & 10 & 20 & 30 & 10 & 20 & 30 \\ \hline
		Accuracy & 98.62 & 98.38 & 97.82 & 97.41 & 96.66 & 94.72 \\ \hline
		CIFAR-100 & \multicolumn{3}{c|}{Symmetry-$50\%$} & \multicolumn{3}{c|}{Pair-$45\%$} \\ \hline
		$T_{update}$ & 10 & 20 & 30 & 10 & 20 & 30 \\ \hline
		Accuracy & 53.84 & 53.61 & 52.87 & 38.57 & 39.88 & 38.67 \\ \hline
	\end{tabular}
	\label{ablation2}
\end{table}
\textbf{Choice of $R(T)$:}
The value of the dropping rate $R(T)$ is set to be $1-\epsilon. \min (\frac{T}{T_k}, 1)$ and common profiles of $R(T)$ for different noise levels $\epsilon$ is shown in Figure \ref{motivation}.
Essentially, after $T \geq T_k$, $R(T) = 1- \epsilon$, which implies that the training tries to concentrate on the correctly labeled portion of the data.
It was observed in~\cite{coteaching} that if too many instances are dropped, then the networks might not get sufficient training data and the performance can substantially deteriorate.
However, in our work, since we are re-labeling the large loss samples and utilizing the possibly correct ones for training, the performance of the proposed approach may not be so intricately related to the value of $\epsilon$.
We analyze the performance of our algorithm in Table \ref{ablation1} by setting $\epsilon=0.5$ to compute $R(T)$.
We choose this value, since, in general, it is difficult to train deep based noise-tolerant models for a larger noise rate ($\epsilon > 0.5$) without any additional information \cite{coteaching}.
From Table \ref{ablation1}, we observe that the performance of mCT-S2R is quite consistent, which indicates that even without the knowledge of the true noise level $\epsilon$, our framework should perform well.

\begin{table}[t!]
	\small
	\centering
	\caption{Average test accuracy (over the last ten epochs) on MNIST, CIFAR-10 and CIFAR-100 datasets for $\epsilon=0.5$ as compared to the recommended value ($\epsilon=$ actual noise level, denoted by ``*") as proposed in \cite{coteaching}.}
	\renewcommand{\arraystretch}{1.1}
	\setlength{\tabcolsep}{2.0 pt}
	\begin{tabular}{|c|c|c|c|c|c|c|}
		\hline
		\multicolumn{1}{|l|}{} & \multicolumn{2}{l|}{MNIST} & \multicolumn{2}{l|}{CIFAR-10} & \multicolumn{2}{l|}{CIFAR-100} \\ \hline
		$\epsilon$ & $*$ & $0.5$ & $*$ & $0.5$ & $*$ & $0.5$ \\ \hline
		Pair-45$\%$ & \begin{tabular}[c]{@{}c@{}}94.72\\ $\pm$ 0.02\end{tabular} & \begin{tabular}[c]{@{}c@{}}97.79\\ $\pm$ 0.02\end{tabular} & \begin{tabular}[c]{@{}c@{}}80.58\\ $\pm$ 0.54\end{tabular} & \begin{tabular}[c]{@{}c@{}}83.53\\ $\pm$ 0.10\end{tabular} & \begin{tabular}[c]{@{}c@{}}38.67\\ $\pm$ 0.08\end{tabular} & \begin{tabular}[c]{@{}c@{}}41.06\\ $\pm$ 0.13\end{tabular} \\ \hline
		Symmetry-20$\%$ & \begin{tabular}[c]{@{}c@{}}98.86\\ $\pm$ 0.01\end{tabular} & \begin{tabular}[c]{@{}c@{}}99.45\\ $\pm$ 0.02\end{tabular} & \begin{tabular}[c]{@{}c@{}}87.21\\ $\pm$ 0.04\end{tabular} & \begin{tabular}[c]{@{}c@{}}83.65\\ $\pm$ 0.07\end{tabular} & \begin{tabular}[c]{@{}c@{}}60.49\\ $\pm$ 0.36\end{tabular} & \begin{tabular}[c]{@{}c@{}}60.76\\ $\pm$ 2.14\end{tabular} \\ \hline
	\end{tabular}
	\label{ablation1}
\end{table}
\textbf{Choice of the softmax threshold value $\kappa$:} 
The parameter $\kappa$ controls which large loss re-labeled samples are picked for further training of the network.
We analyze the performance of our algorithm in Table \ref{ablation3} by setting $\kappa=\{0.80,0.90,0.95\}$ and observe that the results are consistent for a wide range of $\kappa$ values.

\begin{table}[h!]
	\small
	\centering
	\caption{Average test accuracy (over the last ten epochs) on the MNIST and CIFAR-100 datasets for different values of $\kappa$.}
	\renewcommand{\arraystretch}{1.1}
	\setlength{\tabcolsep}{5.0 pt}
	\begin{tabular}{|c|c|c|c|c|c|c|}
		\hline
		MNIST & \multicolumn{3}{c|}{Symmetry-$50\%$} & \multicolumn{3}{c|}{Pair-$45\%$} \\ \hline
		$\kappa$ & 0.80 & 0.90 & 0.95 & 0.80 & 0.90 & 0.95 \\ \hline
		Accuracy & 94.34 & 94.72 & 94.95 & 97.91 & 97.82 & 97.79 \\ \hline
		CIFAR-100 & \multicolumn{3}{c|}{Symmetry-$50\%$} & \multicolumn{3}{c|}{Pair-$45\%$} \\ \hline
		$\kappa$ & 0.80 & 0.90 & 0.95 & 0.80 & 0.90 & 0.95 \\ \hline
		Accuracy & 38.14 & 38.67 & 37.95 & 53.19 & 52.87 & 53.04 \\ \hline
	\end{tabular}
	\label{ablation3}
\end{table}
\textbf{Choice of the network for re-labeling:}
Here, we analyze the performance of our method when one of the networks (in random) is chosen for re-labeling and subsequent training. 
We observe from the results in Table \ref{ablation4} that the performance does not vary much and hence our framework is robust to the choice of the network used for re-labeling.

\begin{table}[h!]
	\small
	\centering
	\caption{Average test accuracy (over the last ten epochs) on the MNIST, CIFAR-10 and CIFAR-100 datasets for different noise levels: (a) Pair-$45\%$, (b) Symmetry-$50\%$ and (c) Symmetry-$20\%$ with the $1^{st}$ (first row) and $2^{nd}$ (second row) network used for re-labeling respectively.}
	\renewcommand{\arraystretch}{1.1}
	\setlength{\tabcolsep}{2.0 pt}
	\begin{tabular}{|c|c|c|c|c|c|c|c|c|c|}
		\hline
		& \multicolumn{3}{c|}{MNIST} & \multicolumn{3}{c|}{CIFAR-10} & \multicolumn{3}{c|}{CIFAR-100} \\ \hline
		& (a) & (b) & (c) & (a) & (b) & (c) & (a) & (b) & (c) \\ \hline
		$1^{st}$ & 94.72 & 97.82 & 98.86 & 80.58 & 81.23 & 87.21 & 38.67 & 52.87 & 60.49 \\ \hline
		$2^{nd}$ & 94.56 & 97.53 & 98.88 & 80.76 & 81.10 & 86.99 & 39.57 & 53.00 & 61.23 \\ \hline
	\end{tabular}
	\label{ablation4}
\end{table}    

\section{Conclusion}
In this work, we present a novel framework mCT-S2R for training neural networks under different amounts of label noise. 
The training paradigm uses the concept of small loss instances along with re-labeling to build noise-resistant classification models. 
In addition, we have shown how self-supervision pre-training can effectively help to boost the performance further.
Our framework works on a wide variety of datasets under different noise models and significantly outperforms the current state-of-the-art, while being computationally lighter.

\bibliography{egbib}
\end{document}